\documentclass[conference]{IEEEtran}
\IEEEoverridecommandlockouts

\usepackage{cite}
\usepackage{amsmath,amssymb,amsfonts}
\usepackage{graphicx}
\usepackage{textcomp}
\usepackage{xcolor}
\usepackage{booktabs}
\usepackage{tabularx}
\usepackage{multirow}
\usepackage{stfloats}

\def\BibTeX{{\rm B\kern-.05em{\sc i\kern-.025em b}\kern-.08em
    T\kern-.1667em\lower.7ex\hbox{E}\kern-.125emX}}

\makeatletter
\newcommand{\linebreakand}{\end{@IEEEauthorhalign}\hfill\mbox{}\par\vspace{1.5em}\mbox{}\hfill\begin{@IEEEauthorhalign}}
\makeatother

\begin{document}

\title{Beyond Risk Prediction: Evidence Grounding and Psychosocial Factor Verification for Explainable Suicide Risk Assessment}

\author{
\IEEEauthorblockN{\makebox[0.31\textwidth][c]{1\textsuperscript{st} Tianle Hu}}
\IEEEauthorblockA{\makebox[0.31\textwidth][c]{\begin{tabular}{@{}c@{}}
\textit{Department of Information, Artificial}\\
\textit{Intelligence and Data Science}\\
\textit{Daiichi Institute of Technology}\\
Tokyo, Japan\\
0009-0007-5929-6546
\end{tabular}}}
\and
\IEEEauthorblockN{\makebox[0.31\textwidth][c]{2\textsuperscript{nd} Chen Peng}}
\IEEEauthorblockA{\makebox[0.31\textwidth][c]{\begin{tabular}{@{}c@{}}
\textit{Department of Information, Artificial}\\
\textit{Intelligence and Data Science}\\
\textit{Daiichi Institute of Technology}\\
Tokyo, Japan\\
0009-0002-6304-0141
\end{tabular}}}
\and
\IEEEauthorblockN{\makebox[0.31\textwidth][c]{3\textsuperscript{rd} Yi-Hsin Tsai}}
\IEEEauthorblockA{\makebox[0.31\textwidth][c]{\begin{tabular}{@{}c@{}}
\textit{Department of Information, Artificial}\\
\textit{Intelligence and Data Science}\\
\textit{Daiichi Institute of Technology}\\
Tokyo, Japan\\
0000-0002-0215-9440
\end{tabular}}}
\linebreakand
\IEEEauthorblockN{\makebox[0.31\textwidth][c]{4\textsuperscript{th} Takshing Andy Tung}}
\IEEEauthorblockA{\makebox[0.31\textwidth][c]{\begin{tabular}{@{}c@{}}
\textit{Research and Development of}\\
\textit{Collaborative Educational Practices}\\
\textit{Graduate School of Education}\\
\textit{Tokyo Gakugei University}\\
Tokyo, Japan\\
0009-0001-3194-1060
\end{tabular}}}
\and
\IEEEauthorblockN{\makebox[0.31\textwidth][c]{5\textsuperscript{th} Bingyang Sun}}
\IEEEauthorblockA{\makebox[0.31\textwidth][c]{\begin{tabular}{@{}c@{}}
\textit{Department of Information, Artificial}\\
\textit{Intelligence and Data Science}\\
\textit{Daiichi Institute of Technology}\\
Tokyo, Japan\\
0009-0008-1661-1187
\end{tabular}}}
\and
\IEEEauthorblockN{\makebox[0.31\textwidth][c]{6\textsuperscript{th} Yenjou Wang\textsuperscript{*}}}
\IEEEauthorblockA{\makebox[0.31\textwidth][c]{\begin{tabular}{@{}c@{}}
\textit{Department of Information, Artificial}\\
\textit{Intelligence and Data Science}\\
\textit{Daiichi Institute of Technology}\\
Tokyo, Japan\\
0000-0002-1319-2190
\end{tabular}}}
}

\maketitle

\begin{abstract}
Identifying suicide risk from social networking services (SNS) posts is important for detecting suicide-related signals in online environments. However, risk classification alone provides limited insight into the textual evidence and psychosocial factors behind a prediction. Based on the IEEE BigData 2026 Explainable Suicide Risk Detection Challenge, this study presents a framework consisting of Risk Assessment, Evidence Grounding, and Factor Identification. Risk Assessment uses length-based routing to accommodate posts of different lengths. Evidence Grounding identifies supporting phrases and uses a Risk--Evidence constraint to maintain consistency with the Risk prediction. For Factor Identification, two verifiers are used. The Taxonomy Verifier focuses on factor semantics, whereas the Evidence-Aware Verifier uses factor-specific lexical-semantic cues to select informative positive training units. Their prediction probabilities are combined to produce the final factor predictions. The three tasks are evaluated using task-specific F1 score measures. Risk Assessment achieved a Weighted F1 of 0.8088, Evidence Grounding achieved a test Macro row F1 of 0.7605, and Factor Identification achieved a Macro F1 of 0.5562. The results show that the framework can provide risk predictions, along with supporting textual evidence and fine-grained information on psychosocial factors. Overall, the proposed framework extends suicide-risk assessment beyond risk-level prediction and provides a more interpretable analysis of SNS posts.
\end{abstract}


\section{Introduction}

Social networking services (SNS) have become important platforms for expressing emotions, life stress, and psychological distress, providing a source of natural-language data for mental-health research in addition to traditional clinical information~\cite{resnik2021}. As such data has become increasingly available, identifying suicidal thoughts and risk indicators from platforms such as Reddit using natural language processing and machine-learning techniques has received growing attention~\cite{yeskuatov2022}. However, suicide-related posts are often complex and may contain multiple psychological, social, behavioral, and protective factors, making reliable and interpretable assessment difficult.

Previous research has mainly formulated suicide-risk detection as a text-classification problem. Pretrained language models, semi-supervised learning, pseudo-labeling, model ensembles, user-level information, and ordinal risk structures have all been explored to improve classification performance~\cite{ruiz2019clpsych,lovitt2024suicide,nguyen2024leveraging,wang2025powerm,yang2025hierarchical}. However, a correct risk prediction alone does not explain which parts of the text support the decision or which psychosocial factors are reflected in the post. Recent studies have therefore begun to investigate textual evidence and fine-grained risk factors~\cite{chen2024,chen2024prompts}. Nevertheless, risk, evidence, and factor predictions are often treated separately, which may lead to outputs that are individually plausible but mutually inconsistent.

The Three-Step Theory (3ST) provides a useful psychological perspective for understanding this limitation. The 3ST distinguishes the formation of suicidal ideation, the intensification of ideation depending on connectedness, and the transition from ideation to action through the capacity for suicide~\cite{klonsky2015three}. This perspective suggests that suicide risk cannot be fully represented by a single risk label because different psychological, social, and behavioral conditions may be involved in suicidal ideation and its transition to action. Therefore, a more explainable assessment should consider not only the risk state, but also the textual evidence supporting the assessment and the psychosocial factors reflected in the post.

Motivated by these limitations, this study presents a factor-centered
framework in which Risk Assessment and Evidence Grounding serve as
supporting components for Psychosocial Factor Identification. The main
contributions are:

\begin{itemize}
    \item A coordinated multi-component framework integrates
        length-aware Risk Assessment, risk-consistent Evidence Grounding,
        and psychosocial Factor Identification.
    \item A dual-verifier Factor Identification method combines
        taxonomy-oriented verification with cue-guided positive-unit selection
        under strict cross-fitted evaluation.
    \item The resulting factor profiles support theory-informed
        interpretation beyond risk-level prediction.
\end{itemize}

\section{Related Work}

\subsection{Social Media-Based Suicide Risk Assessment}

Suicide-risk assessment from social-media text has increasingly been
formulated as a classification problem. Earlier work relied on
linguistic features and conventional classifiers, whereas recent
studies have adopted pretrained language models and more advanced
training strategies. Ruiz et al.~\cite{ruiz2019clpsych} used multiple
Reddit posts from the same user for four-level suicide-risk prediction,
showing the usefulness of cross-post information. Lovitt et
al.~\cite{lovitt2024suicide} addressed limited labeled data and class
imbalance through semi-supervised learning and pseudo-labeling, while
Nguyen and Pham~\cite{nguyen2024leveraging} combined LLM-based
pseudo-labeling with model ensembles.

Recent work has also incorporated information beyond single-post
classification. Wang et al.~\cite{wang2025powerm} introduced temporal
features and ensemble learning, and Yang et
al.~\cite{yang2025hierarchical} modeled ordinal as well as categorical
relationships among risk levels. These studies indicate that contextual
information and structured relationships among risk classes can improve
risk prediction. However, useful context varies across posts, motivating
the length-aware routing adopted in this study.

\subsection{Evidence-Grounded and Psychosocial Factor-Aware Modeling}

A predicted risk label alone provides limited information about the
textual basis of the decision. Accordingly, recent work has explored
evidence and rationale extraction from suicide-related text. The
CLPsych 2024 shared task summarized multiple approaches to identifying
evidence of suicidality~\cite{chim2024overview}. Tran and
Matsui~\cite{tran2024ism} used knowledge self-generation and output
refinement with a large language model, Chen et
al.~\cite{chen2024prompts} investigated instructive prompts for
risk-supporting evidence extraction, Tanaka and
Fukazawa~\cite{tanaka2024integrating} combined supervised extractive
and generative models, and Varadarajan et
al.~\cite{varadarajan2024archetypes} explored theory-driven evidence
extraction. These studies improve interpretability, but risk prediction
and evidence extraction are often modeled separately, which can lead to
inconsistency between the predicted risk state and its supporting
evidence.

Beyond textual evidence, psychosocial factors provide a finer-grained
description of the conditions represented in a post. Li et
al.~\cite{li2025pfa} introduced the Protective Factor-Aware (PFA)
taxonomy containing 19 risk factors and five protective factors, while
Ophir et al.~\cite{ophir2020deep} demonstrated the usefulness of
psychosocial signals for suicide-risk modeling. Factor identification
remains difficult because related labels may have overlapping semantics,
and many factors are expressed only implicitly. This motivates the use of definition-conditioned verification and
cue-guided positive-unit selection in the present framework.

Psychological theories can further support interpretation of
psychosocial factors. The Integrated Motivational--Volitional Model,
3ST, and Interpersonal Theory of Suicide describe
suicidal thoughts and behaviors through constructs related to
hopelessness, connectedness, interpersonal conditions, and capability
~\cite{oconnor2018imv,klonsky2015three,vanorden2010interpersonal}.
In this study, however, 3ST is used only as a theoretical perspective
for interpreting selected group-level factor contrasts; it is not used
as a training constraint or as a direct definition of the computational
factor labels.

Overall, prior work has typically emphasized risk classification,
evidence extraction, or psychosocial factor modeling separately. Our
framework retains Risk Assessment and Evidence Grounding as supporting
components while focusing on definition-conditioned and cue-guided
Psychosocial Factor Identification.

\section{Methodology}

\begin{figure*}[t]
    \centering
    \includegraphics[width=0.95\textwidth]{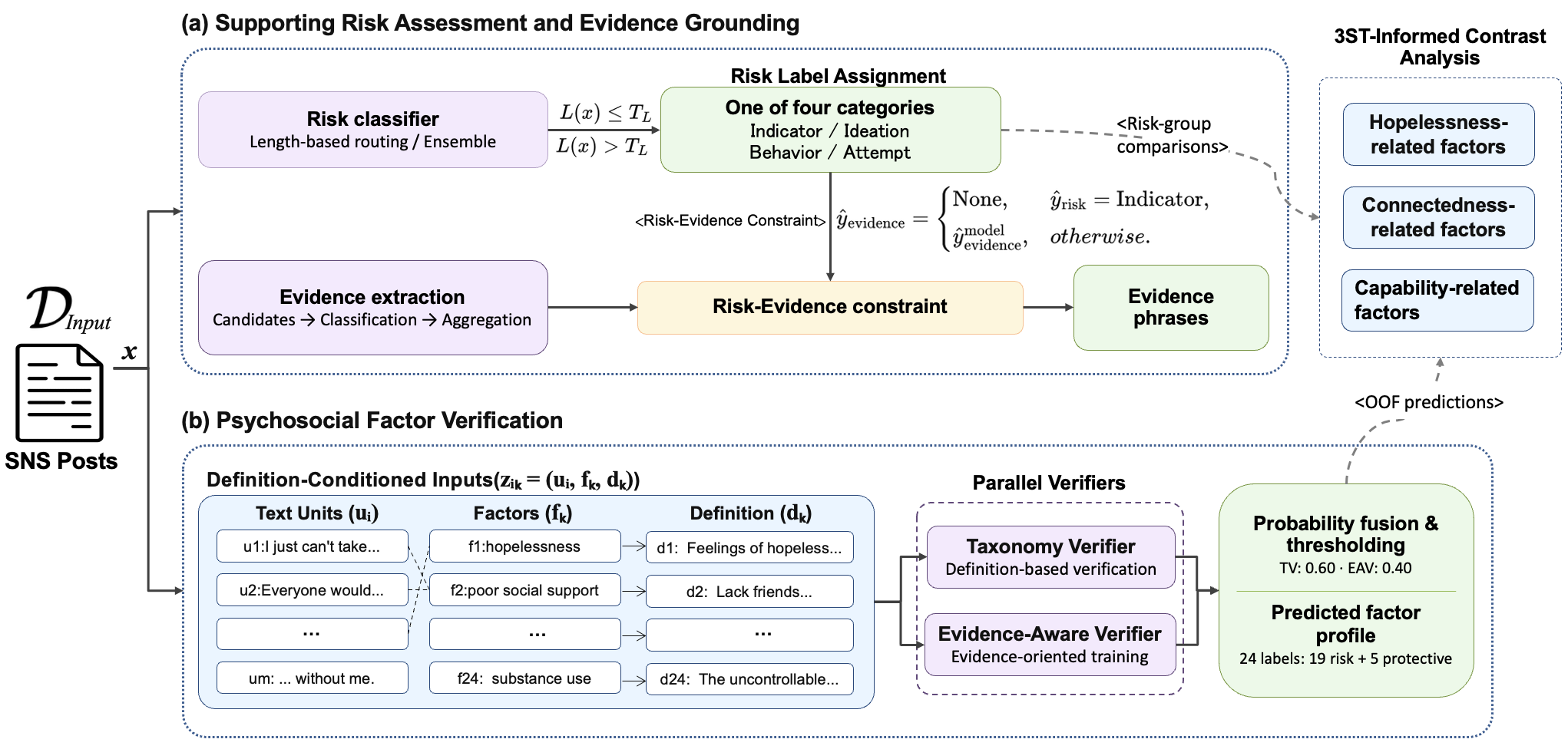}
    \caption{Overview of the proposed factor-centered framework with
supporting Risk Assessment and Evidence Grounding and post-hoc
3ST-informed interpretation.}
    \label{fig:framework}
\end{figure*}

To provide a more comprehensive and explainable assessment of suicide
risk from SNS posts, we develop a coordinated framework comprising
Risk Assessment, Evidence Grounding, and Psychosocial Factor
Identification, as illustrated in Fig.~\ref{fig:framework}. The three
components provide complementary views of risk level, supporting
evidence, and psychosocial factors.

% Framework

\subsection{Supporting Risk Assessment and Evidence Grounding}
\label{sec:supporting_modules}

Risk Assessment and Evidence Grounding are retained as supporting
components of the framework shown in Fig.~\ref{fig:framework}.
Risk Assessment assigns each post to one of four risk categories:
Indicator, Ideation, Behavior, or Attempt~\cite{bigdata2026suicide}.
To accommodate variation in post length, deterministic length-based
routing is used. Let $L(x)$ denote the token length of post $x$ and
$T_L$ the routing threshold. Posts satisfying $L(x)\leq T_L$ are
processed by a Short-Context Ensemble, whereas longer posts are
processed by a Long-Context Specialist. The routed probability vector
is

\begin{equation}
\mathbf{p}_{\mathrm{route}}(x)=
\begin{cases}
\mathbf{p}_{\mathrm{short}}(x), & L(x)\leq T_L,\\
\mathbf{p}_{\mathrm{long}}(x),  & L(x)>T_L,
\end{cases}
\end{equation}

where the short-context prediction is obtained by equal-weight
averaging of the constituent models. The final risk prediction is

\begin{equation}
\widehat{y}_{\mathrm{risk}}(x)
=
\underset{c\in\mathcal{Y}}{\arg\max}\;
p_{\mathrm{route},c}(x).
\end{equation}

The short-context branch is a three-seed equal-probability ensemble of DeBERTa-v3-large models~\cite{he2023debertav3} with a maximum sequence length of 384, using seeds 514, 9, and 42. The long-context branch uses a matched seed-514 DeBERTa-v3-large model with the maximum sequence length extended to 640. Routing is deterministic and based on the full, untruncated token count including special tokens: inputs with \(L(x)\le384\) use the short-context ensemble, whereas longer inputs use the 640-token branch.

Evidence Grounding identifies textual spans that support the risk assessment. Candidate phrases are generated around suicide-related lexical anchors and classified by a DeBERTa-v3-base binary classifier. After evidence-specific annotation cleaning, candidate phrases are constructed from local contexts surrounding the lexical anchors. Hard negative candidates are manually reviewed, and ambiguous cases
are excluded from the hard-negative training set. Candidates whose evidence probabilities exceed a threshold $\beta$ are ranked by probability and aggregated using overlap-aware selection, in which lower-probability overlapping candidates are discarded.

A Risk--Evidence constraint is then applied to prevent an Indicator
prediction from being accompanied by suicide-related evidence:

\begin{equation}
\hat{E}_{\mathrm{final}}(x) =
\begin{cases}
\texttt{none}, &
\hat{y}_{\mathrm{risk}}(x)=\mathrm{Indicator},\\
\hat{E}_{\mathrm{model}}(x), &
\text{otherwise}.
\end{cases}
\end{equation}

\subsection{Psychosocial Factor Identification}
\label{sec:factor_method}

Psychosocial Factor Identification is formulated as a 24-label
multi-label classification problem following the taxonomy introduced
by Li et al.~\cite{li2025pfa}. The taxonomy comprises 19 risk factors
and five protective factors. Since multiple psychosocial factors may be
expressed within a single post, the target output is represented as a
24-dimensional binary vector.

Each post $x$ is segmented into sentence-like text units
$\{u_i\}_{i=1}^{N_x}$. For each factor $f_k$, a semantic definition
$d_k$ is provided according to the adopted taxonomy. Each text unit is
paired with the target factor and its definition to construct a
definition-conditioned input,

\begin{equation}
    \mathbf{z}_{ik}
    =
    (u_i,f_k,d_k),
    \label{eq:factor_input}
\end{equation}

where $u_i$ denotes the $i$th text unit and $d_k$ the definition of
factor $f_k$. This formulation allows verification against factor
semantics rather than relying only on the factor name.

\paragraph{Taxonomy Verifier.}
The Taxonomy Verifier (TV) uses DeBERTa-v3-base
\cite{he2023debertav3} to assess the semantic compatibility between a
text unit and the definition of the target factor. Because each input
contains both the factor identity and its semantic definition, TV
performs factor verification against the adopted taxonomy rather than
relying only on lexical overlap with the factor name.

\paragraph{Evidence-Aware Verifier.}
The Evidence-Aware Verifier (EAV) uses the same
definition-conditioned formulation and model backbone as TV, but
differs in the construction of positive training units. For each factor, a manually specified lexical-semantic cue list is used
in a rule-based scoring function to rank sentence-like units. Cue
matches within the unit receive the largest score contribution, while
matches elsewhere in the post receive a smaller contribution; a small
number of factor-specific heuristic bonuses and penalties further refine
the ranking. Up to three positive units with scores of at least 1.0 are
selected for training; if none reaches this threshold, the highest-
scoring unit is retained as a fallback. These cues are used only for
training-unit selection and are independent of the output of the
Evidence Grounding module.

Both verifiers produce unit-level factor probabilities and use the
same max-pooling rule for post-level aggregation. For verifier
$v\in\{\mathrm{TV},\mathrm{EAV}\}$,

\begin{equation}
    p_k^{(v)}(x)
    =
    \max_i p_{ik}^{(v)},
    \qquad
    v\in\{\mathrm{TV},\mathrm{EAV}\}.
    \label{eq:factor_pooling}
\end{equation}

Thus, the distinction between TV and EAV lies in the construction of
their positive training units rather than in the post-level aggregation
rule. Max pooling allows a strongly supporting text unit to activate the
corresponding factor at the post level.

\paragraph{Probability-Level Fusion.}
TV and EAV therefore provide two views of the same factor-verification
task: TV emphasizes compatibility with the factor taxonomy, whereas
EAV emphasizes units containing more direct factor-related support.
Their post-level probabilities are combined as

\begin{equation}
    p_k^{\mathrm{final}}(x)
    =
    \alpha p_k^{\mathrm{TV}}(x)
    +(1-\alpha)p_k^{\mathrm{EAV}}(x),
    \label{eq:factor_fusion}
\end{equation}

where $p_k^{\mathrm{TV}}(x)$ and $p_k^{\mathrm{EAV}}(x)$ denote the
post-level probabilities produced by TV and EAV, respectively. In the
final system, $\alpha=0.60$, assigning weights of 0.60 and 0.40 to TV
and EAV.

A factor is activated according to its decision threshold $\tau_k$:

\begin{equation}
    \hat{y}_k(x)
    =
    \mathbb{I}\left(
    p_k^{\mathrm{final}}(x)\geq\tau_k
    \right),
    \qquad k=1,\ldots,24,
    \label{eq:factor_decision}
\end{equation}

where $\mathbb{I}(\cdot)$ is the indicator function. Under the strict
cross-fitted evaluation protocol, the threshold used for a target fold
is determined without using that fold. A global threshold floor of
0.50 is applied, while a conservative floor of 0.60 is used for
meaning in life, sense of responsibility, exposure to others' suicide,
poor school performance, sexual orientation-related issues, cognitive
deficits, and substance use.

The final multi-label prediction is therefore

\begin{equation}
    \hat{\mathbf{y}}_{\mathrm{factors}}
    =
    [\hat{y}_1,\ldots,\hat{y}_{24}]
    \in \{0,1\}^{24},
    \label{eq:factor_prediction}
\end{equation}

where each element indicates whether the corresponding psychosocial
factor is activated for post $x$.

The predicted psychosocial factor profile is further examined from a
theory-informed perspective. Selected factors are interpreted with
reference to 3ST, focusing on themes related to hopelessness,
connectedness, and capability. This interpretation is applied only as
a post-hoc analytical perspective and is not used as a training
constraint or as a direct definition of the computational factor
labels. The detailed training configuration and cross-fitted evaluation
protocol are described in Section~\ref{sec:factor_results}.

% =========================================================
% IV. EXPERIMENTS AND RESULTS
% =========================================================

\section{Experiments and Results}

\begin{table*}[t]
\caption{Context-Length and Routing Ablation for Risk Assessment}
\label{tab:risk_ablation}
\centering
\footnotesize
\setlength{\tabcolsep}{8pt}
\renewcommand{\arraystretch}{1.08}

\begin{tabular*}{\textwidth}{@{\extracolsep{\fill}}llcccc@{}}
\toprule
\textbf{Ablation} &
\textbf{Setting} &
\textbf{Weighted F1} &
\textbf{Macro F1} &
\textbf{Attempt Recall} &
\textbf{Behavior Recall} \\
\midrule

Context length (matched single seed) &
ML384 &
0.7897 &
0.7631 &
0.5877 &
\textbf{0.7724} \\

&
ML640 &
\textbf{0.8051} &
\textbf{0.7857} &
\textbf{0.6930} &
0.7519 \\[1mm]

Routing (three-seed ML384 baseline) &
ML384 AVG3 &
0.8067 &
0.7802 &
0.6228 &
\textbf{0.7673} \\

&
Routed ML384/ML640 &
\textbf{0.8088} &
\textbf{0.7841} &
\textbf{0.6404} &
0.7621 \\

\bottomrule
\end{tabular*}
\end{table*}

\subsection{Dataset and Evaluation Protocol}
\label{sec:dataset_protocol}

This study uses the dataset released for the IEEE BigData 2026
Explainable Suicide Risk Detection Challenge
\cite{bigdata2026suicide,li2025pfa}. The challenge is designed around
two official subtasks. The first requires prediction of suicide-risk
level together with supporting textual evidence, while the second
requires identification of psychosocial factors associated with the
post. To reflect these requirements in a modular form, we organize the
system into three components: Risk Assessment, Evidence Grounding, and
Psychosocial Factor Identification.

The labeled corpus contains 1,635 posts from 153 unique users. An
additional 378 unlabeled samples were provided for leaderboard
evaluation and final submission, but their gold labels are not publicly
available. Therefore, the analyses reported in this paper are based
primarily on the labeled corpus.

Because the three components correspond to different prediction units
and evaluation settings, task-specific protocols are used. Risk
Assessment is evaluated using five-fold
\texttt{StratifiedGroupKFold} cross-validation with
\texttt{anon\_user\_id} as the grouping variable and split seed 514,
ensuring that posts from the same user do not appear simultaneously in
the training and held-out subsets of a fold.

Psychosocial Factor Identification is also evaluated using a
user-grouped five-fold protocol with split seed 42. For each target
fold, the held-out posts are predicted only by models that were not
trained on those posts, producing pooled out-of-fold predictions over
the labeled corpus. Factor-specific thresholds for a target fold are
determined from the remaining folds, forming a strict cross-fitted
evaluation procedure.

Evidence Grounding uses a separate post-level split because its target
is a textual span within each individual post rather than a
user-level trajectory. After evidence-specific data cleaning, 1,611
posts are divided into training, validation, and test subsets using a
70/10/20 split based on \texttt{row\_id}, preventing the same post
from appearing in multiple subsets. Because Evidence Grounding is
evaluated independently at the post level without using user identity
or cross-post context, this split does not introduce direct sample
overlap.

The official primary metrics are Weighted F1 for Risk Assessment,
Phrase F1 for Evidence Grounding, and Macro F1 for Factor
Identification. Because the three components use different prediction
targets and evaluation protocols, their scores are interpreted
task-specifically. Row-based Macro F1 is additionally used for
Evidence Grounding model and threshold selection. Experiments were
conducted on NVIDIA T4, RTX 3090, RTX A6000, and RTX PRO 6000
Blackwell GPUs.

\subsection{Supporting Risk and Evidence Results}
\label{sec:supporting_results}

The final length-routed Risk Assessment system uses 384- and 640-token
short- and long-context branches with $T_L=384$, achieving an
out-of-fold (OOF) Weighted F1 of 0.8088.

To disentangle the effects of context length and deterministic routing,
we conduct two complementary ablations, as summarized in
Table~\ref{tab:risk_ablation}. For context length, ML384 and ML640 are
compared under a matched single-seed setting. Increasing the context
length to 640 improves Weighted F1 from 0.7897 to 0.8051, Macro F1 from
0.7631 to 0.7857, and Attempt recall from 0.5877 to 0.6930, while
Behavior recall decreases from 0.7724 to 0.7519. For routing, the routed ML384/ML640 system is compared against the
canonical three-seed ML384 ensemble. On the OOF evaluation, the routed
configuration obtains a Weighted F1 of 0.8088 compared with 0.8067 for
the ML384 ensemble, with corresponding Macro F1 values of 0.7841 and
0.7802. Attempt recall increases from 0.6228 to 0.6404, whereas
Behavior recall decreases from 0.7673 to 0.7621. However, the routed
configuration did not outperform the ML384 ensemble in standalone Risk
performance on the blind leaderboard. Because Risk and Evidence are
coupled in the first challenge subtask, we therefore additionally
examine Indicator-specific behavior, since an Indicator prediction
directly suppresses downstream evidence through the Risk--Evidence
constraint.

Among the 1,635 OOF samples, 43 are assigned to the long-context
branch. Routing changes 17 argmax predictions, including 9 error
corrections and 6 newly introduced errors, yielding a net gain of
three correct predictions. Indicator-specific analysis further shows that the routed configuration
reduces false Indicator predictions from 42 to 37 relative to the
ML384 three-seed ensemble, while the number of true Indicator
predictions changes only from 514 to 513. Indicator precision therefore
increases from 0.9245 to 0.9327, with nearly unchanged recall
(0.8412 vs.\ 0.8396). Among the six samples for which the two systems
differ in Indicator assignment, five are false Indicator predictions
made only by the ML384 ensemble, including four Ideation posts and one
Attempt post.

This distinction is particularly relevant to the joint Risk--Evidence
pipeline because a false Indicator prediction forces the corresponding
evidence output to \texttt{none}, potentially suppressing valid
risk-supporting evidence.

For Evidence Grounding, 1,611 usable posts remain after
evidence-specific cleaning. We remove 23 Indicator posts with
non-\texttt{none} evidence annotations and one full-post annotation,
and revise 28 paraphrased or mismatched annotations when the
corresponding source span can be identified. The classifier uses
DeBERTa-v3-base with a maximum input length of 256, a learning rate of
$1\times10^{-5}$, weight decay of 0.01, and up to five training epochs
with seed 42. The decision threshold is set to $\beta=0.50$ using
validation Macro F1. After overlap-aware aggregation and application of
the Risk--Evidence constraint, the held-out test Macro row F1 is
0.7605.

\begin{table*}[t]
\caption{Overall Factor Identification Results}
\label{tab:factor_overall}
\centering
\scriptsize
\setlength{\tabcolsep}{5pt}
\renewcommand{\arraystretch}{0.95}
\begin{tabular}{@{}lcccccc@{}}
\toprule
\textbf{Model} &
\textbf{Macro F1} &
\textbf{Micro F1} &
\textbf{Samples F1} &
\textbf{Weighted F1} &
\textbf{Exact Match} &
\textbf{Avg. Labels/Post} \\
\midrule
TV
& .5467 & .6551 & .5640 & .6609 & .1865 & 3.554 \\
EAV
& .5288 & .6294 & .5640 & .6425 & .1376 & 4.113 \\
\textbf{Fusion}
& \textbf{.5562}
& \textbf{.6635}
& \textbf{.5783}
& \textbf{.6689}
& \textbf{.1890}
& \textbf{3.489} \\
\bottomrule
\end{tabular}

\vspace{1mm}
\footnotesize{Gold average label cardinality: 2.919.}
\end{table*}

\begin{table*}[!b]
\caption{Factor Taxonomy and Per-Factor F1 Scores}
\label{tab:factor_per_label}
\centering
\scriptsize
\setlength{\tabcolsep}{2.5pt}
\renewcommand{\arraystretch}{0.90}

\begin{tabularx}{0.96\textwidth}{
@{}
>{\raggedright\arraybackslash}X
ccc
@{\hspace{12pt}}
>{\raggedright\arraybackslash}X
ccc
@{}
}
\toprule
\textbf{Factor} & \textbf{TV} & \textbf{EAV} & \textbf{Fusion} &
\textbf{Factor} & \textbf{TV} & \textbf{EAV} & \textbf{Fusion} \\
\midrule

Hopelessness
& .767 & .770 & \textbf{.773} &
Low socio-economic status (LSS)
& .695 & .638 & \textbf{.710} \\

Low self-esteem (LS)
& .743 & \textbf{.774} & .753 &
Traumatic experience (TE)
& .387 & \textbf{.462} & .394 \\

Emotion dysregulation (ED)
& \textbf{.692} & .592 & .687 &
Cognitive deficits
& \textbf{.462} & .442 & .459 \\

Poor social support (PSS)
& \textbf{.756} & .733 & .754 &
Substance use
& .529 & .514 & \textbf{.545} \\

Suicide means (SM)
& .721 & .713 & \textbf{.729} &
Exposure to others' suicide
& \textbf{.364} & .263 & .323 \\

Interpersonal difficulty (ID)
& .533 & .510 & \textbf{.560} &
Poor school performance
& .458 & \textbf{.512} & .489 \\

Mental health issues (MHI)
& .699 & .698 & \textbf{.702} &
Sexual orientation-related issues (SORI)
& \textbf{.471} & .381 & \textbf{.471} \\

Prior self-harm or suicidal thought/attempt (PSST)
& .568 & .563 & \textbf{.591} &
Coping strategy (CS)$^\dagger$
& .713 & .714 & \textbf{.733} \\

Stressful life event (SLE)
& \textbf{.550} & .485 & .541 &
Psychological capital (PC)$^\dagger$
& \textbf{.521} & .474 & .505 \\

Dysfunctional family (DF)
& .681 & .682 & \textbf{.760} &
Social support (SS)$^\dagger$
& .447 & .380 & \textbf{.479} \\

Interpersonal violence (IV)
& .475 & .450 & \textbf{.480} &
Sense of responsibility$^\dagger$
& .309 & .318 & \textbf{.324} \\

Physical health (PH)
& .447 & .454 & \textbf{.457} &
Meaning in life$^\dagger$
& .133 & \textbf{.173} & .130 \\

\bottomrule
\end{tabularx}

\vspace{1mm}
\footnotesize{$^\dagger$ Protective factor; all others are risk factors.}
\end{table*}

\subsection{Factor Identification Results}
\label{sec:factor_results}

Psychosocial Factor Identification is evaluated as a 24-label
post-level multi-label classification task over all 1,635 labeled
posts. TV and EAV are both implemented using DeBERTa-v3-base and
trained independently under a five-fold protocol. The maximum input
length is 320 tokens, with a batch size of 48 and five training epochs.
AdamW~\cite{loshchilov2019adamw} is used with encoder and
classification-head learning rates of $1\times10^{-5}$ and
$2\times10^{-4}$, respectively, weight decay of 0.01, and a linear
learning-rate schedule with an 8\% warmup ratio. Binary cross-entropy
with logits and fold-specific positive-class weighting is used for
training.

For strict cross-fitted evaluation, each post is evaluated only by the
verifier model for which that post is held out. Factor-specific
decision thresholds for a target fold are determined from the
remaining four folds. TV and EAV probabilities are then combined using
fixed weights of 0.60 and 0.40, respectively. A global threshold floor
of 0.50 is applied, with a conservative floor of 0.60 for meaning in
life, sense of responsibility, exposure to others' suicide, poor
school performance, SORI, cognitive deficits, and substance use. The
TV and EAV fusion weights were fixed at 0.60 and 0.40, respectively,
and the conservative 0.60 threshold floor for seven factors was
inherited from the competition-development configuration rather than
tuned on the strict cross-fitted predictions reported here.

Table~\ref{tab:factor_overall} shows that TV is stronger than EAV
standalone, while Fusion achieves the best performance across all four
aggregate F1 measures, improving Macro F1 from 0.5467 to 0.5562.
Fusion also reduces average predicted label cardinality from 3.554 for
TV and 4.113 for EAV to 3.489, compared with 2.919 in the gold labels.

Across the five held-out folds, mean Macro F1 $\pm$ standard deviation
was $0.5435\pm0.0332$ for TV and $0.5527\pm0.0304$ for Fusion.
Fusion improved over TV in four of the five folds; the mean
fold-level Fusion--TV difference was $+0.0092\pm0.0206$, indicating
a modest but non-uniform gain across folds.

A post-hoc sensitivity analysis under the same strict cross-fitted
protocol showed that Macro F1 remained between 0.5562 and 0.5600
across the tested TV weights from 0.45 to 0.75. The highest value in
this range was 0.5600 at $\alpha=0.50$, consistent with the fact that
the adopted $\alpha=0.60$ was not selected by optimizing the reported
OOF results and that performance was relatively insensitive to moderate
changes in the fusion weight.

Table~\ref{tab:factor_per_label} further shows that the relative
strengths of TV and EAV vary across factors. Fusion achieves the
highest F1 on 13 factors and ties the highest value for SORI. Relative
to TV, Fusion improves F1 on 16 of the 24 factors.

The standalone verifiers also show different factor-level strengths.
EAV performs best for low self-esteem, traumatic experience, poor
school performance, and meaning in life, whereas TV performs best for
emotion dysregulation, poor social support, stressful life event,
cognitive deficits, exposure to others' suicide, and psychological
capital. These differences indicate that TV and EAV provide
non-identical decision signals. Although EAV is weaker in aggregate as
a standalone verifier, its factor-dependent strengths can still
contribute positively when combined with TV.

\begin{table*}[t]
\caption{Selected Risk-Group Factor Contrasts in Gold Annotations and Fusion Predictions}
\label{tab:risk_factor_contrasts}
\centering
\scriptsize
\setlength{\tabcolsep}{3.5pt}
\renewcommand{\arraystretch}{0.92}

\begin{tabularx}{\textwidth}{@{}
>{\raggedright\arraybackslash}X
>{\raggedright\arraybackslash}X
c
c
c
c
@{}}
\toprule
\textbf{Factor} &
\textbf{Contrast} &
\textbf{Gold $\Delta$} &
\textbf{Fusion $\Delta$} &
\textbf{Fusion 95\% CI} &
\textbf{$p$} \\
\midrule

Hopelessness
& Ideation--Indicator
& +.165 & +.169 & [.119, .219] & $<$.001 \\

Poor social support
& Ideation--Indicator
& -.060 & -.105 & [-.171, -.039] & $<$.001 \\

Interpersonal difficulty
& Ideation--Indicator
& +.005 & +.029 & [-.017, .078] & .227 \\

Dysfunctional family
& Ideation--Indicator
& +.048 & +.050 & [.010, .090] & .015 \\

Suicide means
& (Behavior+Attempt)--Ideation
& +.477 & +.400 & [.344, .457] & $<$.001 \\

PSST
& (Behavior+Attempt)--Ideation
& +.252 & +.214 & [.159, .273] & $<$.001 \\

Suicide means
& Attempt--Ideation
& +.294 & +.365 & [.282, .461] & $<$.001 \\

PSST
& Attempt--Ideation
& +.809 & +.506 & [.400, .601] & $<$.001 \\

\bottomrule
\end{tabularx}
\end{table*}

\subsection{Exploratory Risk-Factor Distributional Analysis}
\label{sec:risk_factor_analysis}

To examine whether the predicted factor profiles preserve selected
risk-group patterns observed in the gold annotations, we conduct a
post-hoc distributional analysis from a 3ST-informed perspective.
Contrasts are computed using the fixed OOF Fusion predictions without
retraining the factor models. Uncertainty is estimated using 5,000 user-level cluster bootstrap
resamples of the fixed OOF predictions. For each bootstrap sample,
users are resampled with replacement and the corresponding between-group
contrast is recomputed. The 95\% confidence interval is obtained from
the 2.5th and 97.5th percentiles of the bootstrap contrast distribution.
Two-sided $p$-values are estimated as twice the smaller proportion of
bootstrap contrasts falling on either side of zero. Thus, the reported
uncertainty reflects variation from resampling the observed users rather
than model-training uncertainty.

Factor selection is theory-driven rather than performance-driven. We
focus on factors related to three themes considered in 3ST:
hopelessness, connectedness-related conditions, and capability-related
conditions. These factor labels are treated as empirical proxies for
broader theoretical themes rather than direct measurements of the
psychological constructs.

For the Ideation--Indicator comparison,
Table~\ref{tab:risk_factor_contrasts} shows close agreement for
hopelessness, with a gold contrast of $+.165$ and a Fusion contrast of
$+.169$. Dysfunctional family also shows similar differences
($+.048$ and $+.050$, respectively). Poor social support exhibits a
negative contrast in both the gold annotations and Fusion predictions,
with a larger magnitude in the predictions. Interpersonal difficulty
shows only a small difference, and its confidence interval includes
zero, indicating substantial uncertainty for this contrast.

For contrasts involving suicidal behavior and attempt, suicide means
and prior self-harm or suicidal thought/attempt (PSST) show positive
differences relative to Ideation in both the gold annotations and
Fusion predictions. For suicide means, the
(Behavior+Attempt)--Ideation contrast is $+.477$ in the gold
annotations and $+.400$ in Fusion. For PSST, the corresponding values
are $+.252$ and $+.214$. The largest discrepancy appears in the
Attempt--Ideation comparison for PSST, where the gold contrast is
$+.809$ while Fusion produces $+.506$, suggesting that the model
preserves the direction of the group difference but compresses its
magnitude.

Overall, the selected Fusion contrasts generally follow the directions
observed in the gold annotations, but the degree of agreement varies
across factors. These findings describe empirical group-level patterns
under a 3ST-informed lens and should not be interpreted as validation
of 3ST, direct measurement of its psychological constructs, or
evidence of longitudinal progression.

\section{Discussion}

The Risk ablation shows that additional context is useful but does not
benefit all classes uniformly. In the matched comparison, ML640 improves
the aggregate F1 measures and Attempt recall, but reduces Behavior
recall. The effect of deterministic routing is also small and depends
on the evaluation setting: the routed configuration provides a slight
OOF improvement but does not outperform the ML384 ensemble in
standalone Risk performance on the blind leaderboard.

For the joint Risk--Evidence pipeline, however, the type of Risk error
is also important. Because an Indicator prediction suppresses all
downstream evidence, false Indicator predictions can directly reduce
evidence recall. The routed configuration reduces such false Indicator
predictions with almost no change in Indicator recall. These results
therefore suggest that the Risk component should be examined not only
through aggregate classification scores, but also through its
interaction with downstream Evidence Grounding.

The factor-level results suggest that TV and EAV emphasize different
aspects of psychosocial factor expression. TV directly evaluates semantic compatibility with factor definitions,
whereas EAV uses factor-specific lexical-semantic cues to bias positive
training-unit selection toward more directly supportive units. Their fusion therefore combines
two verifiers with different inductive biases rather than simply
averaging two similarly behaving models. This helps explain why EAV can
remain weaker as a standalone verifier while still contributing useful
factor-specific signals when combined with TV.

The results also highlight the difficulty of fine-grained psychosocial
factor identification. Several factors are expressed indirectly, occur
in relatively few posts, or overlap semantically with related factors.
These properties can make the boundaries between related factors
difficult to distinguish and may contribute to both over-prediction and
uneven per-factor performance. The higher prediction cardinality of EAV is consistent with a broader
factor-activation tendency, while fusion partially controls this
behavior through the stronger taxonomy-oriented signal from TV.

The exploratory risk--factor analysis provides a complementary
group-level view of the predicted factor profiles. The close agreement
observed for some contrasts, such as hopelessness and dysfunctional
family, indicates that Fusion can preserve selected differences observed
in the gold annotations. At the same time, discrepancies such as the
attenuated PSST Attempt--Ideation contrast show that directional
agreement does not imply accurate recovery of contrast magnitude.
Accordingly, the 3ST perspective is used here only to organize and
interpret selected empirical patterns. The analysis does not establish
that the computational factors directly measure 3ST constructs, validate
the theory, or represent longitudinal progression from ideation to
suicidal action.

\subsection{Limitations}

Several limitations remain. The effect of deterministic length routing
is small, only 43 of the 1,635 OOF samples are assigned to the
long-context branch, and its standalone Risk advantage does not
consistently transfer across evaluation settings.  Factor labels are substantially imbalanced,
especially for rare and implicit factors. EAV relies on manually specified lexical-semantic cues for positive-unit
selection, but the quality and coverage of this cue-guided selection
have not been evaluated independently. The analysis is post-level and cannot model
longitudinal changes within users. In addition, the 3ST analysis uses
selected factor labels as proxies for broader theoretical themes rather
than direct psychological constructs.

\section{Conclusion}

Risk-level prediction alone provides limited insight into the textual
evidence and psychosocial factors underlying suicide-related SNS posts.
To address this limitation, this study presented a factor-centered
framework in which Risk Assessment and Evidence Grounding serve as
supporting components for Psychosocial Factor Identification. Fusion of taxonomy-oriented verification and cue-guided
factor-verification signals achieved a Macro F1 of 0.5562, outperforming either verifier alone, while Risk Assessment achieved an OOF Weighted F1 of 0.8088 using the
length-aware configuration, with Indicator-specific analysis showing
the importance of upstream Risk errors for downstream Evidence
Grounding. Evidence Grounding achieved a held-out test Macro row F1 of 0.7605.

These results indicate that combining complementary factor-verification
signals can provide a more fine-grained and interpretable assessment
beyond risk-level prediction alone. Future work will focus on improving rare-factor identification,
systematically evaluating the coverage and quality of cue-guided
positive-unit selection, and extending the framework to longitudinal
user-level analysis.

\section*{Acknowledgment}

We would like to thank Jie-Han Chen of Taipei, Taiwan, for his valuable suggestions on the research direction.

\bibliographystyle{IEEEtran}
\bibliography{references}

\begin{thebibliography}{10}
\providecommand{\url}[1]{#1}
\csname url@samestyle\endcsname
\providecommand{\newblock}{\relax}
\providecommand{\bibinfo}[2]{#2}
\providecommand{\BIBentrySTDinterwordspacing}{\spaceskip=0pt\relax}
\providecommand{\BIBentryALTinterwordstretchfactor}{4}
\providecommand{\BIBentryALTinterwordspacing}{\spaceskip=\fontdimen2\font plus
\BIBentryALTinterwordstretchfactor\fontdimen3\font minus
  \fontdimen4\font\relax}
\providecommand{\BIBforeignlanguage}[2]{{%
\expandafter\ifx\csname l@#1\endcsname\relax
\typeout{** WARNING: IEEEtran.bst: No hyphenation pattern has been}%
\typeout{** loaded for the language `#1'. Using the pattern for}%
\typeout{** the default language instead.}%
\else
\language=\csname l@#1\endcsname
\fi
#2}}
\providecommand{\BIBdecl}{\relax}
\BIBdecl

\bibitem{resnik2021}
P.~Resnik, A.~Foreman, M.~Kuchuk, K.~Musacchio~Schafer, and B.~Pinkham,
  ``Naturally occurring language as a source of evidence in suicide
  prevention,'' \emph{Suicide and Life-Threatening Behavior}, vol.~51, no.~1,
  pp. 88--96, 2021.

\bibitem{yeskuatov2022}
E.~Yeskuatov, S.-L. Chua, and L.~K. Foo, ``Leveraging reddit for suicidal
  ideation detection: A review of machine learning and natural language
  processing techniques,'' \emph{International Journal of Environmental
  Research and Public Health}, vol.~19, no.~16, p. 10347, 2022.

\bibitem{ruiz2019clpsych}
V.~Ruiz, L.~Shi, W.~Quan, N.~Ryan, C.~Biernesser, D.~Brent, and R.~Tsui,
  ``{CLPsych2019} shared task: Predicting suicide risk level from {Reddit}
  posts on multiple forums,'' in \emph{Proceedings of the Sixth Workshop on
  Computational Linguistics and Clinical Psychology}, 2019, pp. 162--166.

\bibitem{lovitt2024suicide}
M.~Lovitt, H.~Ma, S.~Wang, and Y.~Peng, ``Suicide risk assessment on social
  media with semi-supervised learning,'' in \emph{Proceedings of the 2024 IEEE
  International Conference on Big Data (BigData)}, 2024, pp. 8541--8549.

\bibitem{nguyen2024leveraging}
V.~Nguyen and C.~Pham, ``Leveraging large language models for suicide detection
  on social media with limited labels,'' in \emph{Proceedings of the 2024 IEEE
  International Conference on Big Data (BigData)}, 2024, pp. 8550--8559.

\bibitem{wang2025powerm}
L.-F. Wang, P.-C. Chuang, M.-H. Li, and C.-T. Li, ``{PoWER-M}: Prediction of
  writing-based emotional risk with mental-adaptive multi-modal learning,'' in
  \emph{Proceedings of the 2025 IEEE International Conference on Big Data
  (BigData)}, 2025, pp. 8181--8190.

\bibitem{yang2025hierarchical}
C.~Yang, Z.~Wang, W.~Tan, Z.~Tan, C.~Ji, and Z.~Zhou, ``Hierarchical dual-head
  model for suicide risk assessment via {MentalRoBERTa},'' in \emph{Proceedings
  of the 2025 IEEE International Conference on Big Data (BigData)}, 2025, pp.
  8191--8199.

\bibitem{chen2024}
N.~Chen, Z.~Dai, and Y.~Wang, ``Stress and hope distinguish individuals with
  suicidal plan from suicide ideators among chinese college students,''
  \emph{Frontiers in Psychiatry}, vol.~15, p. 1387868, 2024.

\bibitem{chen2024prompts}
J.~Chen, V.~Nguyen, X.~Dai, D.~Molla-Aliod, C.~Paris, and S.~Karimi,
  ``Exploring instructive prompts for large language models in the extraction
  of evidence for supporting assigned suicidal risk levels,'' in
  \emph{Proceedings of the Ninth Workshop on Computational Linguistics and
  Clinical Psychology (CLPsych 2024)}, 2024, pp. 197--202.

\bibitem{klonsky2015three}
E.~D. Klonsky and A.~M. May, ``The three-step theory ({3ST}): A new theory of
  suicide rooted in the ``ideation-to-action'' framework,'' \emph{International
  Journal of Cognitive Therapy}, vol.~8, no.~2, pp. 114--129, 2015.

\bibitem{chim2024overview}
J.~Chim, A.~Tsakalidis, D.~Gkoumas, D.~Atzil-Slonim, Y.~Ophir, A.~Zirikly,
  P.~Resnik, and M.~Liakata, ``Overview of the {CLPsych} 2024 shared task:
  Leveraging large language models to identify evidence of suicidality risk in
  online posts,'' in \emph{Proceedings of the Ninth Workshop on Computational
  Linguistics and Clinical Psychology (CLPsych 2024)}, 2024, pp. 177--190.

\bibitem{tran2024ism}
V.~Tran and T.~Matsui, ``Team {ISM} at {CLPsych} 2024: Extracting evidence of
  suicide risk from {Reddit} posts with knowledge self-generation and output
  refinement using a large language model,'' in \emph{Proceedings of the Ninth
  Workshop on Computational Linguistics and Clinical Psychology (CLPsych
  2024)}, 2024, pp. 191--196.

\bibitem{tanaka2024integrating}
R.~Tanaka and Y.~Fukazawa, ``Integrating supervised extractive and generative
  language models for suicide risk evidence summarization,'' in
  \emph{Proceedings of the Ninth Workshop on Computational Linguistics and
  Clinical Psychology (CLPsych 2024)}, 2024, pp. 270--277.

\bibitem{varadarajan2024archetypes}
V.~Varadarajan, A.~Lahnala, A.~V. Ganesan \emph{et~al.}, ``Archetypes and
  entropy: Theory-driven extraction of evidence for suicide risk,'' in
  \emph{Proceedings of the Ninth Workshop on Computational Linguistics and
  Clinical Psychology (CLPsych 2024)}, 2024, pp. 278--291.

\bibitem{li2025pfa}
J.~Li, X.~Wang, H.~Li, Y.~Yan, H.~V. Leong, L.~Feng, N.~X. Yu, and Q.~Li,
  ``Protective factor-aware dynamic influence learning for suicide risk
  prediction on social media,'' \emph{arXiv preprint arXiv:2507.10008}, 2025.

\bibitem{ophir2020deep}
Y.~Ophir, R.~Tikochinski, C.~S.~C. Asterhan, I.~Sisso, and R.~Reichart, ``Deep
  neural networks detect suicide risk from textual {Facebook} posts,''
  \emph{Scientific Reports}, vol.~10, p. 16685, 2020.

\bibitem{oconnor2018imv}
R.~C. O'Connor and O.~J. Kirtley, ``The integrated motivational--volitional
  model of suicidal behaviour,'' \emph{Philosophical Transactions of the Royal
  Society B: Biological Sciences}, vol. 373, no. 1754, p. 20170268, 2018.

\bibitem{vanorden2010interpersonal}
K.~A. Van~Orden, T.~K. Witte, K.~C. Cukrowicz, S.~R. Braithwaite, E.~A. Selby,
  and T.~E. Joiner, ``The interpersonal theory of suicide,''
  \emph{Psychological Review}, vol. 117, no.~2, pp. 575--600, 2010.

\bibitem{bigdata2026suicide}
{IEEE Big Data Cup 2026}, ``Explainable suicide risk detection,'' Official
  competition website, 2026, [Online]. Available:
  https://www.bigdatacompetition.cn/.

\bibitem{he2023debertav3}
P.~He, J.~Gao, and W.~Chen, ``{DeBERTaV3}: Improving {DeBERTa} using
  {ELECTRA}-style pre-training with gradient-disentangled embedding sharing,''
  in \emph{Proceedings of the Eleventh International Conference on Learning
  Representations}, 2023.

\bibitem{loshchilov2019adamw}
I.~Loshchilov and F.~Hutter, ``Decoupled weight decay regularization,'' in
  \emph{Proceedings of the International Conference on Learning
  Representations}, 2019.

\end{thebibliography}

\end{document}